\documentclass[sigconf]{acmart}
\AtBeginDocument{%
  }

\setcopyright{acmlicensed}
\copyrightyear{2018}
\acmYear{2018}
\acmDOI{XXXXXXX.XXXXXXX}
\acmConference[Conference acronym 'XX]{}{June 03--05,
  2018}{Woodstock, NY}
\acmISBN{978-1-4503-XXXX-X/2018/06}

\usepackage[utf8]{inputenc} 
\usepackage[T1]{fontenc}    
\usepackage{hyperref}       
\usepackage{url}            
\usepackage{booktabs}       

\usepackage{amsfonts}       
\usepackage{nicefrac}       
\usepackage{microtype}      
\usepackage{xcolor}         
\usepackage{array}    
\usepackage{multirow}
\usepackage{graphicx} 
\usepackage{amsmath}
\usepackage{wrapfig}
\usepackage{siunitx}
\usepackage{threeparttable}
\usepackage{caption}
\usepackage{subcaption}
\usepackage{pifont,amssymb}
\usepackage{bm} 
\usepackage{xcolor}
\usepackage{balance}

\definecolor{scene}{RGB}{0, 112, 192}
\definecolor{atmosphere}{RGB}{237, 125, 49}
\definecolor{theme}{RGB}{84, 130, 53}

\begin{document}

\title{LoVR: A Benchmark for Long Video Retrieval in Multimodal Contexts}




\author{Qifeng Cai$^*$}
\affiliation{%
  \institution{East China Normal University}
  \city{Shanghai}
  \country{China}}
\email{qfcai@stu.ecnu.edu.cn}

\author{Hao Liang$^*$}
\affiliation{%
  \institution{Peking University \& Zhongguancun Academy}
  \city{Beijing}
  \country{China}}
\email{hao.liang@stu.pku.edu.cn}

\author{Zhaoyang Han$^*$}
\affiliation{%
  \institution{Huazhong University of Science and Technology}
  \city{Wuhan}
  \country{China}}
\email{zyhan04@hust.edu.cn}

\author{Hejun Dong}
\affiliation{%
  \institution{Beihang University}
  \city{Beijing}
  \country{China}}
\email{jhdonghj@buaa.edu.cn}

\author{Meiyi Qiang}
\affiliation{%
  \institution{Peking University}
  \city{Beijing}
  \country{China}}
\email{myqiang25@stu.pku.edu.cn	}

\author{Ruichuan An}
\affiliation{%
  \institution{Peking University}
  \city{Beijing}
  \country{China}}
\email{arctanxarc@gmail.com}

\author{Quanqing Xu$^\dagger$}
\affiliation{%
  \institution{OceanBase, Ant Group}
  \city{Beijing}
  \country{China}}
\email{xuquanqing.xqq@oceanbase.com}

\author{Bin Cui}
\affiliation{%
  \institution{Peking University}
  \city{Beijing}
  \country{China}}
\email{bin.cui@pku.edu.cn}

\author{Wentao Zhang$^\dagger$}
\affiliation{%
  \institution{Peking University \& Zhongguancun Academy}
  \city{Beijing}
  \country{China}}
\email{wentao.zhang@pku.edu.cn}



\thanks{%
$*$Equal contribution. \quad 
$\dagger$Corresponding author. 
}

\renewcommand{\shortauthors}{Hao Liang et al.}

\begin{abstract}
Long videos contain a vast amount of information, making video-text retrieval an essential and challenging task in multimodal learning and web-scale search. On today’s Web, where users increasingly expect to locate not only relevant pages but also specific long videos or fine-grained clips, existing benchmarks fall short due to limited video duration, low-quality captions, and coarse annotation granularity.
To address these limitations, we introduce \textbf{LoVR}, a benchmark specifically designed for long video-text retrieval. \textbf{LoVR} contains 467 long videos and over 40,804 fine-grained clips with high-quality captions.
To overcome the issue of poor machine-generated annotations, we propose an efficient caption generation framework that integrates VLM automatic generation, caption quality scoring, and dynamic refinement. This pipeline improves annotation accuracy while maintaining scalability. Furthermore, we introduce a semantic fusion method to generate coherent full-video captions without losing important contextual information.
Our benchmark introduces longer videos, more detailed captions, and a larger-scale dataset, presenting new challenges for video understanding and retrieval. Extensive experiments on various advanced models demonstrate that \textbf{LoVR} is a challenging benchmark, revealing the limitations of current approaches and providing valuable insights for future research. We release the code link at 
\url{https://lovrbench.github.io/}
\end{abstract}




\maketitle

\section{Introduction}

\begin{table}[t] 
\centering
\caption{Comparison of statistics of video-text benchmarks.}
\resizebox{\columnwidth}{!}{ 
\begin{tabular}{lcccc}
\toprule
\textbf{Benchmark} & \textbf{\#Sample} & \textbf{Avg. Len.(s)} & \textbf{Avg. Captions} & \textbf{Annotator} \\ 
\midrule
MSR-VTT~\cite{xu2016msr} & 1,000 & 15.01 & 9.41 & Human \\
DiDeMo~\cite{anne2017localizing} & 1,037 & 53.94 & 29.11 & Human \\
MSVD~\cite{chen2011collecting} & 670 & 10.04 & 7.01 & Human \\
ActivityNet~\cite{yu2019activitynet} & 5,044 & 36.00 & 13.48 & Human \\
DREAM-1K~\cite{wang2024tarsier} & 1,000 & 8.90 & 59.30 & Human \\
VDC~\cite{chai2024auroracap} & 1,000 & 28.18 & 500.91 & VLM \\
CAREBENCH~\cite{xu2024fine} & 1,000 & 14.35 & 227.95 & Human \\
\midrule
\multicolumn{5}{c}{\textbf{LoVR}} \\ 
\midrule
\textbf{Clip-level} & 40,804 & 17.86 & 1393.65 & VLM + Human \\
\textbf{Video-level} & 467 & 1560.30 & 106075.39 & VLM + Human \\
\bottomrule
\end{tabular}
}
\label{table:benchmark-comparison}
\end{table}

With the explosive growth of online media platforms, video has become a dominant content form on the Web. Notably, long-form videos, such as lectures, documentaries, and live streams lasting from minutes to hours, constitute a significant and growing portion of this information ecosystem. Effective retrieval of such content is therefore essential for modern web search, where users increasingly expect to locate not just relevant videos but specific moments within them. In this context, video-text retrieval has emerged as a core challenge in multimodal understanding and web-scale search systems~\cite{zhu2023deep}.
\begin{figure*}
  \includegraphics[width=\textwidth]{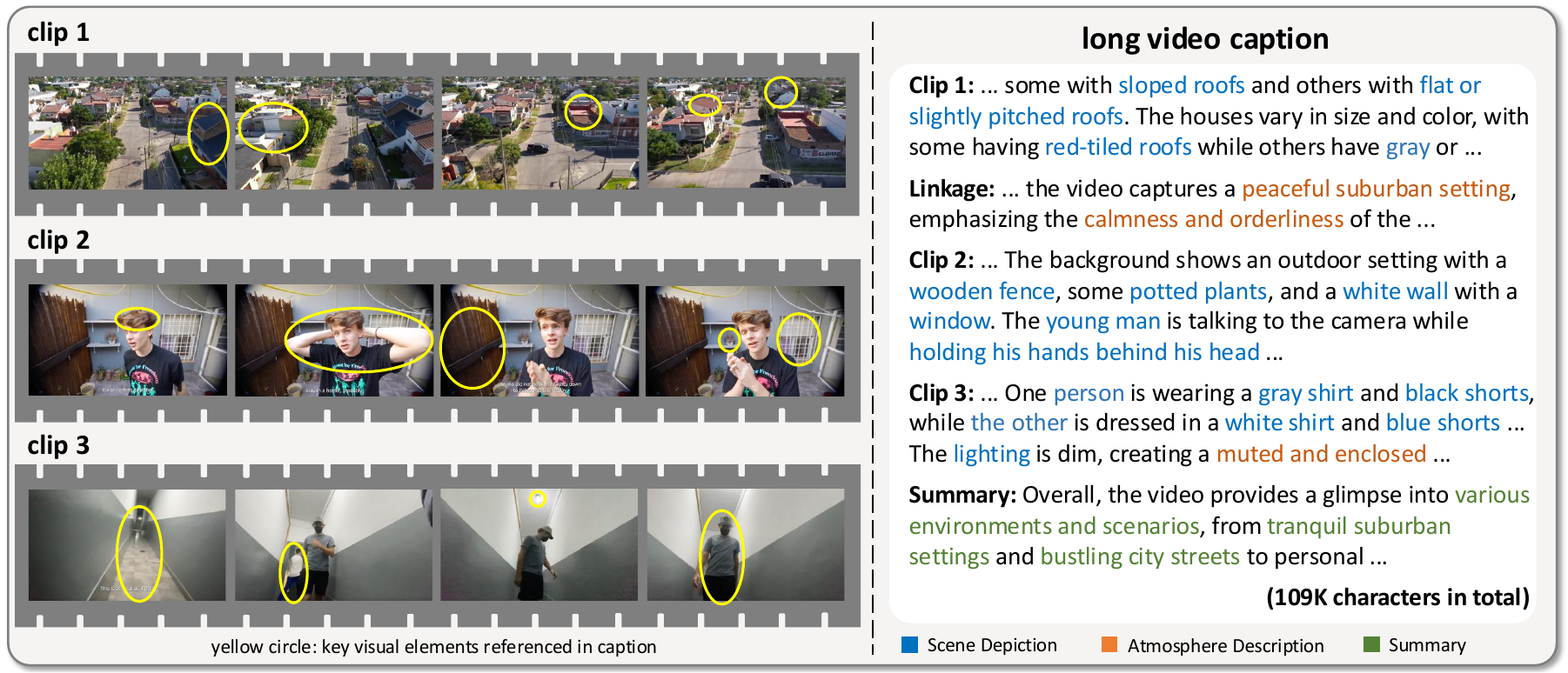}
  \caption{Example from \textbf{LoVR}. A long video is divided into clips, three shown on the left. The caption on the right describes each clip’s \textcolor{scene}{scene} and \textcolor{atmosphere}{atmosphere}, their relations, and the overall \textcolor{theme}{summary}.}
  \label{fig:face}
\end{figure*}
However, as videos grow in length and complexity, supporting retrieval at both the full-video and fine-grained clip level poses a pressing challenge for real-world applications~\cite{lin2022eclipse, luo2024video, eltahir2025multimodal}. This challenge is exacerbated by the fact that current video-text retrieval benchmarks (e.g., MSR-VTT\cite{xu2016msr}, DiDeMo\cite{anne2017localizing}) are primarily designed around short video clips and often feature captions that lack the temporal specificity needed for clip-level localization. Consequently, there is an urgent need for a dedicated benchmark to drive progress in long-form video and clip-level retrieval. The limitations of existing benchmarks can be summarized in three key aspects:

\textbf{(1) Limited Video Length} Most widely used datasets—such as MSR-VTT~\cite{xu2016msr}, MSVD~\cite{chen2011collecting}, and DiDeMo~\cite{anne2017localizing} contain only short videos, with average durations less than 1 minute. These short durations make it difficult to simulate the complexity of real-world long-video retrieval tasks, limiting the ability of such benchmarks to evaluate models on long-form content effectively.

\textbf{(2) Low Video Caption Quality} The annotation quality of captions significantly affects benchmark performance. While many datasets rely on machine-generated annotations (e.g., using Automatic Speech Recognition(ASR)~\cite{malik2021automatic}), these methods often suffer from low accuracy and semantic ambiguity. Although manual annotation~\cite{wu2024longvideobench, song2024moviechat, yu2019activitynet} can improve semantic fidelity, it typically lacks fine-grained detail and is not scalable to large-scale datasets. Balancing scalability with annotation quality remains a critical challenge in current research.

\textbf{(3) Lack of Clip Level Benchmarks} Most existing benchmarks either focus solely on full-video retrieval or rely on pre-segmented short clips that do not originate from continuous long videos~\cite{eltahir2025multimodal, zhu2024video}. This limits their ability to evaluate models' fine-grained temporal localization and contextual understanding within long-form content. In contrast, constructing clip-level benchmarks by segmenting long videos introduces a more realistic and challenging setting that models must not only retrieve relevant content but also distinguish fine-grained clips from densely packed temporal contexts. This setting better reflects practical applications and poses a greater challenge for retrieval models.
\begin{figure*}[t]
    \centering
    \includegraphics[width=\linewidth]{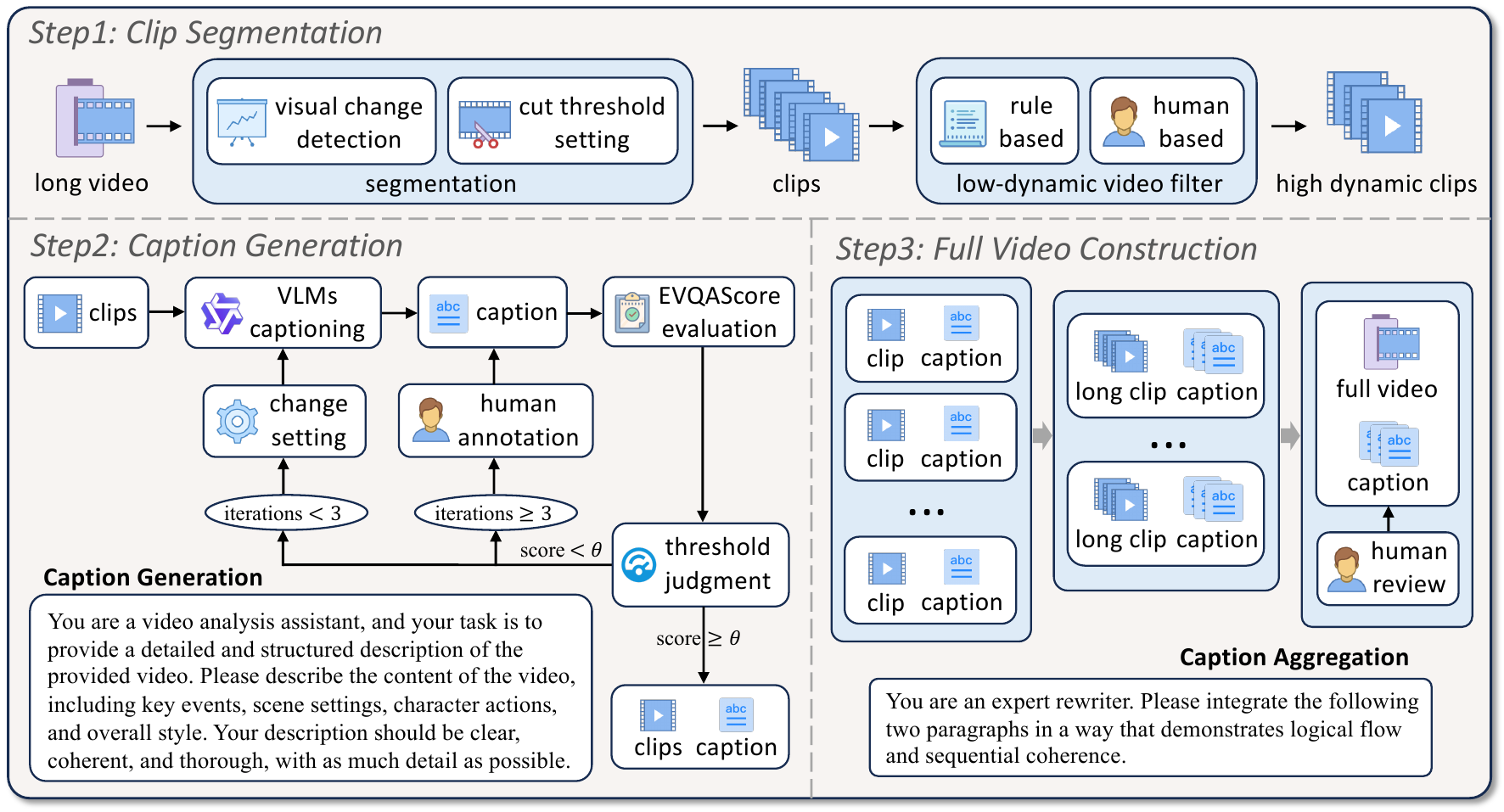}
    \caption{Overview of the data construction pipeline in \textbf{LoVR}. 
    Step 1 segments long videos into high-dynamic clips using visual change detection and threshold-based filtering. Step 2 generates high-quality clip-level captions via iterative VLM captioning and human fallback based on EVQAScore. Step 3 constructs long-video captions by clustering clip captions and summarizing them into a full-length description. Finally, a human review process is conducted to ensure quality.
    }
    \label{fig:dataset_pipeline}
\end{figure*}
To address the aforementioned limitations, we introduce \textbf{LoVR}, a new benchmark specifically designed for long video-text retrieval. Unlike existing datasets that predominantly consist of short video content, \textbf{LoVR} targets the challenges inherent to long-form videos and aims to support both full-video and fine-grained clip-level retrieval. As shown in Figure \ref{fig:face}, the caption of the full video not only contains holistic information but also corresponds to fine-grained clip-level details.

To overcome the shortage of high-quality annotations at scale, we design a cost-effective and scalable caption generation pipeline that balances annotation quality with efficiency. Our pipeline comprises three key stages: (1) generating initial captions using a state-of-the-art Vision-Language Model (VLM)~\cite{zhang2024vision}, (2) automatically assessing caption quality via the EVQAScore~\cite{liang2024evqascore}, and (3) conducting a final round of human verification. This hybrid process effectively ensures both high fidelity and scalability of annotations. For caption generation, we adopt the advanced Qwen2.5-VL-Instruct model~\cite{wang2024qwen2, bai2025qwen2}, which demonstrates strong performance on multimodal understanding tasks. The integration of EVQAScore~\cite{liang2024evqascore} further enhances the reliability of our automatic validation step by offering fine-grained quality assessments aligned with human judgments. By ensuring annotation quality while maintaining scalability, \textbf{LoVR} offers a more realistic and comprehensive evaluation benchmark that better reflects the demands of real-world long video retrieval scenarios. 

Evaluation on the \textbf{LoVR} benchmark reveals several fundamental challenges faced by current video-text retrieval methods in long video scenarios. 
First, the extended temporal span and structural complexity of long videos make it difficult for models to effectively capture and represent all relevant semantic content~\cite{ge2024v2pe, zhao2023mmicl}. As video length increases, both the computational cost of feature extraction and index construction rise significantly, while meaningful semantic signals become sparser.
Second, captions play a vital role in aligning textual and visual modalities. In the context of long videos, captions are typically longer and semantically richer, which increases the difficulty of achieving precise video-text alignment~\cite{chen2025expertized}. Accurately modeling high-level semantics, such as overarching themes, narrative progression, and emotional tone, remains an open challenge~\cite{faghihi2024spectrum}.

These observations underscore the importance of a dedicated benchmark. By explicitly targeting long-form content, \textbf{LoVR} surfaces technical bottlenecks in current approaches and provides a valuable foundation for developing more robust long video-text retrieval systems. Our contributions are as follows:

\begin{enumerate}
    \item We introduce \textbf{LoVR}, a novel benchmark designed for long video-text retrieval. It comprises 467 full-length videos tailored for long video retrieval, along with 40,804 fine-grained clips to support clip-level retrieval within long videos. Each video and clip is paired with high-quality captions that have undergone both human and machine validation. To the best of our knowledge, \textbf{LoVR} is the first large-scale dataset dedicated to advancing research in long-form video retrieval. Table \ref{table:benchmark-comparison} compares \textbf{LoVR} and other existing benchmarks. 
    \item To ensure caption quality, we propose a general and cost-effective caption generation pipeline that integrates VLMs with both automated and human validation. Compared to fully automatic or fully manual annotation, our approach achieves high-quality captions while significantly reducing human labor by effectively balancing machine efficiency and human oversight.

    \item We conduct systematic experiments using various models on \textbf{LoVR}, providing insights into effective representation learning and retrieval strategies for long-form video-text understanding. Our results reveal that \textbf{LoVR} is more challenging than existing benchmarks, especially in the context of long video retrieval, thus introducing a new challenge and setting a fresh research direction for the community.
\end{enumerate}

\begin{figure*}[t]
    \centering
    \includegraphics[width=\linewidth]{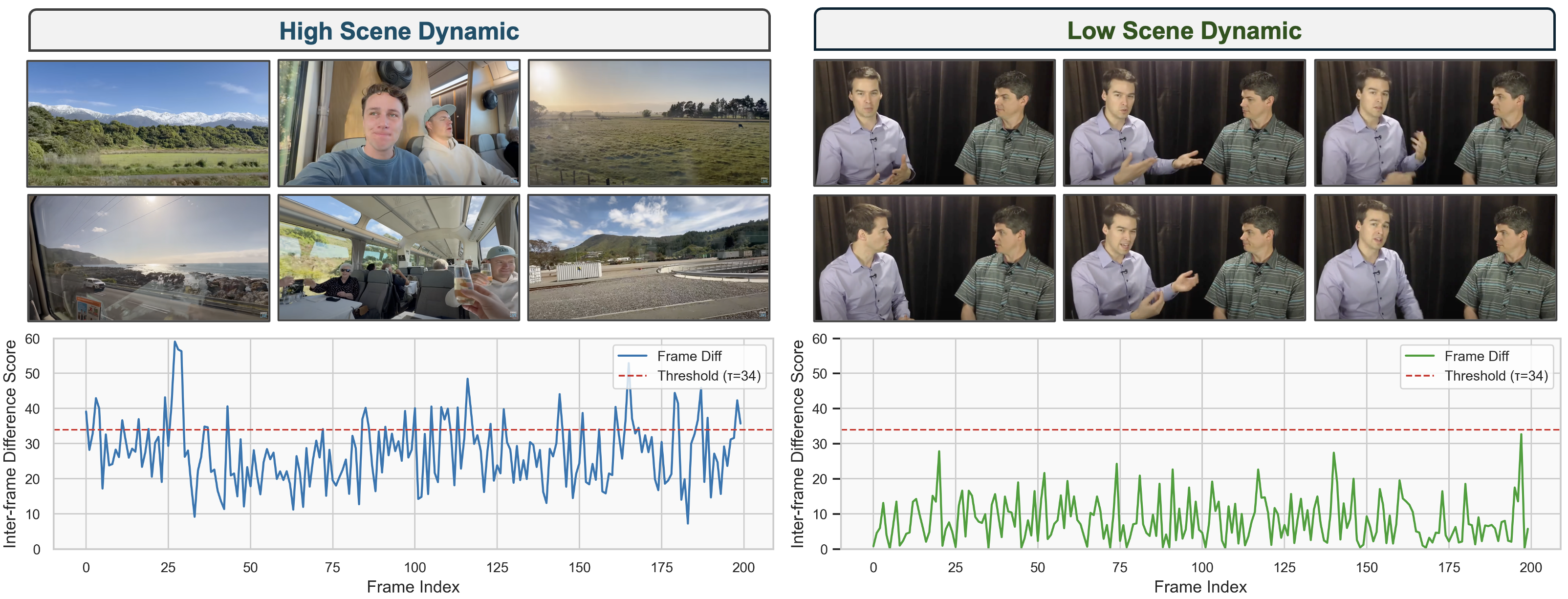}
    \caption{Illustration of Video Dynamics: High vs. Low Scene Dynamics. Example video frames are shown in the upper part (high-dynamic on the left, low-dynamic on the right), while the lower part presents the computed inter-frame difference curves. The high-dynamic video exhibits large frame-to-frame variations, whereas the low-dynamic video remains visually stable. }
    \label{fig:video_type_comparison}
\end{figure*}
\section{Related Works}

\paragraph{Video-text retrieval benchmarks} Traditional benchmarks such as MSR-VTT~\cite{xu2016msr}, MSVD~\cite{chen2011collecting}, DiDeMo~\cite{anne2017localizing} TGIF~\cite{tgif-cvpr2016} and LSMDC~\cite{lsmdc} focus on short video clips paired with concise captions, which limits their capacity to evaluate fine-grained semantic understanding. 
Recent efforts~\cite{yu2019activitynet} aim to scale both the temporal extent of videos and the descriptive richness of annotations, among which HowTo100M~\cite{miech2019howto100m} provides a massive-scale dataset of long-duration tutorial videos, but relies on ASR-generated captions, using only audio to generate labels introduces significant label noise.
To address this limitation, DREAM-1K~\cite{wang2024tarsier} and VDC~\cite{chai2024auroracap} adopt human-annotated descriptions to improve annotation accuracy.
However, their heavy reliance on manual annotation constrains scalable dataset expansion, as the labor-intensive process complicates large-scale data acquisition.

\paragraph{Video-text retrieval methods} Recent advances have been driven by deep learning techniques. For feature extraction, Transformer-based architectures like ViT and CLIP now dominate, effectively capturing spatiotemporal video features and multimodal data (e.g., audio, OCR), while pretrained language models (e.g., BERT) provide superior text representations.~\cite{zhu2023deeplearningvideotextretrieval} Feature matching has evolved from global(e.g. CLIP4Clip~\cite{luo2021clip4clipempiricalstudyclip},VSE++~\cite{han2022bicnetlearningefficientspatiotemporal}) to multi-granular approaches (local and individual-level alignment)(e.g. CLIP2TV~\cite{gao2022clip2tvalignmatchdistill}), improving retrieval accuracy. 
However, challenges remain in cross-modal alignment and computational efficiency, pointing to future research directions in model optimization and weakly supervised learning.



\section{LoVR Datasets Construction}

The \textbf{LoVR} dataset was constructed through a rigorous process involving raw video selection(Section~\ref{sec:video_source}) and dataset construction pipeline(Section~\ref{sec:dataset_pipeline}). Section~\ref{sec:dataset_statistics} provides an overview of \textbf{LoVR} and highlights its key advantages in comparison to existing benchmarks.

\subsection{Video Source}
\label{sec:video_source}
The \textbf{LoVR} benchmark is constructed based on the LongVideoBench~\cite{wu2024longvideobench} dataset. LongVideoBench contains a diverse collection of videos with durations ranging from 8 seconds to 1 hour. The domains of the videos can be broadly categorized into four domains: Life, Movie, Knowledge, and News. To focus on long-form video retrieval, we apply a duration filter and retain only those videos exceeding 15 minutes (900 seconds) in length.

\subsection{Dataset Construction Pipeline}
\label{sec:dataset_pipeline}

As illustrated in Figure~\ref{fig:dataset_pipeline}, the dataset construction pipeline proposed in this paper consists of three key stages: (1) \textit{clip segmentation}, (2) \textit{caption generation}, and (3) \textit{full video captioning}. This pipeline is designed to enhance both the scalability and the overall quality of the benchmark. Following these stages, \textit{final human quality assessment} is conducted to ensure the reliability and quality of the benchmark.

\begin{figure*}[t]
    \centering
    \includegraphics[width=\linewidth]{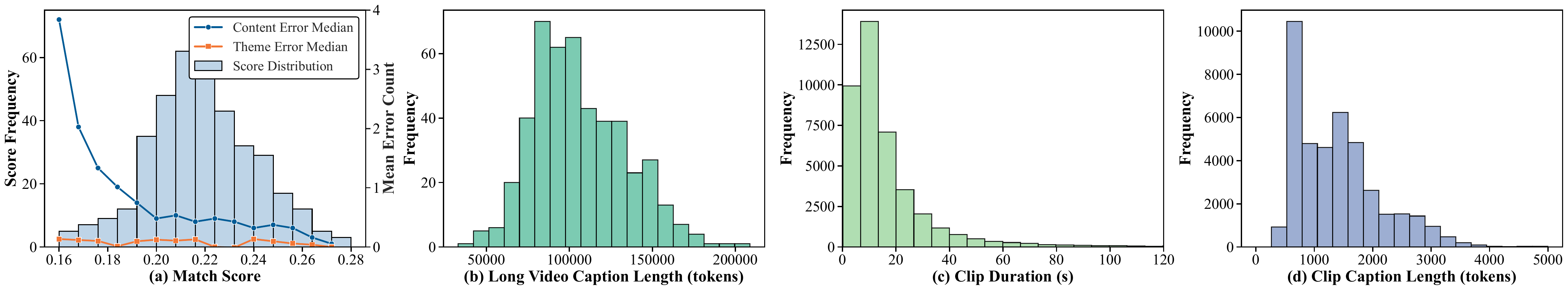}
    \caption{The figure shows four panels from left to right: (a) Distribution of match scores and annotation errors on the sampled subset $K'$. The left y-axis shows the match score distribution, while the right y-axis displays the median content and theme annotation errors. A notable reduction in annotation errors occurs when the match score exceeds 0.2. (b) Distribution of caption lengths for long videos, with most captions concentrated around 10,000 tokens. (c) Distribution of clip durations. (d) Distribution of caption lengths for video clips.}
\label{fig:clip_match_distribution}
\end{figure*}
\subsubsection{Clip Segmentation}
\label{sec:video_filtering}

Clips serve as fundamental semantic units of videos. Each long video $ v \in \mathcal{V} $ is composed of a sequence of temporally contiguous clips, denoted as $ v = \{ c_1, c_2, \dots, c_N \} $, where $ c_i \in \mathcal{C} $. Here, $ \mathcal{V} $ denotes the set of all long videos, and $ \mathcal{C} $ represents the collection of all clips. Segmenting long videos into shorter clips facilitates more fine-grained content analysis and retrieval.
We employ the \texttt{ContentDetector} module from \texttt{PySceneDetect}~\cite{PySceneDetect}, which segments videos based on visual content changes across frames.
This approach relies on a threshold $\tau$ to detect scene changes: when the inter-frame difference surpasses $\tau$, a cut is generated.

Based on visual scene dynamics, videos can be naturally categorized into two types: those with high scene dynamics and those with low scene dynamics.
As illustrated in Figure~\ref{fig:video_type_comparison}, videos such as lectures, interviews, or static photography exhibit minimal visual variation throughout their duration, with scenes remaining largely unchanged across different segments. Therefore, we set $\tau=34$, segmenting clips with high-scene dynamics while preserving those with low-scene dynamics.

The inclusion of low-scene dynamic videos simplifies the retrieval task, as only a small number of frames may suffice to represent the entire content.
In contrast, videos with high scene dynamics contain rich and diverse visual transitions, requiring more precise matching and understanding, which is precisely the level of challenge our benchmark aims to introduce.
Furthermore, as shown in the lower portion of Figure~\ref{fig:video_type_comparison}, the inter-frame difference scores of two representative videos are computed and visualized. The high-scene-dynamics video shows large frame-wise variations, while the low-dynamic one maintains consistently high similarity across frames, often below the threshold for clip segmentation.

To ensure sufficient visual dynamics of the videos, we apply a filtering criterion based on the number of clips generated after segmentation.
Specifically, videos yielding fewer than 40 clips are excluded, as this indicates insufficient visual variation in certain segments.
These filtered videos often contain numerous long-duration clips with little to no change in content, thereby reducing overall scene dynamics.
Finally, we conduct a manual inspection to evaluate the efficacy of our thresholding strategy and confirm its validity.
Ultimately, we curated a final set of 467 high-quality long videos that exhibit rich scene dynamics and semantic content diversity, from which a total of 40,804 clips were extracted.

\subsubsection{Caption Generation}
\label{caption_generation}
We propose a three-stage annotation framework: (1) \textit{automatic captioning}, (2) \textit{quality assessment}, and (3) \textit{dynamic optimization}. This approach integrates VLM-based automatic captioning with human verification, which enables scalable yet precise annotation of clip-level contents.

\paragraph{Automatic Captioning.}
We utilize advanced VLMs to produce initial captions based on the visual content of each clip. Specifically, for each clip $ c_i \in v$, its caption is generated as $\text{cap}_i = \text{VLM}(c_i)$
where $ \text{VLM} $ denotes the process of caption generation using VLM. To ensure high caption quality, we employ Qwen2.5-VL-7B-Instruct~\cite{bai2025qwen2}, a series of advanced VLMs. When generating captions, the VLMs are prompted to describe key aspects of the given clip, including events, scenes, character actions, and overall visual style. 


\paragraph{Quality Assessment.}
To assess the quality of generated captions, we utilize \text{EVQAScore}~\cite{liang2024evqascore}, a fine-grained method designed to measure the semantic alignment between video clips and their corresponding captions.
For each clip-caption pair $ (c_i, \text{cap}_i) $, we compute a matching score $ s_i = \text{EVQAScore}(c_i, \text{cap}_i) $, which quantifies the semantic relevance between the visual content and the generated caption.
After scoring all instances, we construct a dataset $ K = \{(c_i, s_i, \text{cap}_i)\}_{i=1}^n$, containing the video clip, its associated caption, and the corresponding quality score. 

However, there exists a discrepancy between the computed $ s_i $ and whether the caption meets human-annotated quality standards.
To bridge this gap, we determine an empirical threshold $ \theta $ that serves as a benchmark for identifying captions that satisfy manual verification criteria. Specifically, we sample a subset $ K' \subseteq K $ and conduct manual annotation to count captioning errors per instance. This allows us to establish an empirical correlation between the computed matching scores and human-verified relevance.
As illustrated in Figure~\ref{fig:clip_match_distribution}(a), a significant rise in annotation errors is observed for samples with scores below 0.2. Based on this analysis, we set the quality threshold to $ \theta = 0.2 $. Captions with scores above $ \theta $ are retained as high-quality, while those below are flagged as low-quality, needing refinement or exclusion.

\begin{table*}[t]
\caption{Baseline performance of text-to-video and text-to-clip retrieval on the \textbf{LoVR} benchmark.}
\centering
\resizebox{\textwidth}{!}{
\begin{tabular}{l|ccc|rr|ccc|rr}
\toprule
\multirow{2}{*}{\textbf{Model}} & \multicolumn{5}{c|}{\textbf{Text-to-Video Retrieval}} & \multicolumn{5}{c}{\textbf{Text-to-Clip Retrieval}} \\ 
\cmidrule{2-6} \cmidrule{7-11}
 & \textbf{R@1} & \textbf{R@5} & \textbf{R@10} & \textbf{Time w/ IO} & \textbf{Time w/o IO} & \textbf{R@1} & \textbf{R@5} & \textbf{R@10} & \textbf{Time w/ IO} & \textbf{Time w/o IO} \\
\midrule

\multicolumn{11}{c}{\textbf{Image Encoder}} \\ \midrule
CLIP& 23.34& 43.90& 54.39& 2544747& 1431709& 18.78& 37.28& 46.31& 222270298& 125052299\\
MetaCLIP-ViT-H-14 & 26.76& 53.10& 65.52& 1165648& 313637& 25.26& 46.10& 55.09& 101813266& 27394590\\
siglip-base-patch16-224 & 19.06& 37.90& 52.25& 1376746& 344674& 19.34& 36.81& 45.26& 120251547& 30105548\\
siglip2-base-patch16-224 & 10.06& 27.41& 37.04& 1336867& 269687& 8.35& 19.03& 25.28& 116768361& 23555767\\
EVA02-CLIP-B-16 & 26.34& 51.61& 62.96& 7491183& 5156938& 23.37& 44.16& 53.54& 654315581& 450431558\\
PE-Core-B16-224 & 23.34& 48.60& 58.24& 7460584& 4967311& 19.95& 38.42& 47.30& 651642897& 433868639\\
\midrule

\multicolumn{11}{c}{\textbf{Video Encoder}} \\ \midrule
VideoClip-XL & 30.84& 54.82& 65.74& 1074042& 29435& 47.33& 72.66& 80.81& 93812003& 2571057\\
VideoClip-XL-v2 & 29.98& 55.67& 67.67& 1073304& 29497& \textbf{55.34}& \textbf{78.29}& \textbf{84.95}& 93747515& 2576415\\
\midrule

\multicolumn{11}{c}{\textbf{Multimodal Embedding Model}} \\ \midrule
GME-Qwen2-VL & 19.91& 41.54& 53.32& 19242308& 18301540& 53.53& 75.89& 82.72& 1680714676& 1598543574\\
MM-Embed& 14.13& 31.05& 40.69& 2030535& 1168405& 32.32& 52.84& 60.96& 177356626& 102054123\\
LanguageBind-Video& \textbf{42.61}& \textbf{66.60}& \textbf{77.94}& 180170& 180170& 40.30& 65.13& 74.26& 15742371& 15742370\\
\bottomrule
\end{tabular}
}
\label{table:retrieval_table1_new}
\end{table*}

\paragraph{Dynamic Optimization.}
For captions identified as low-quality, we switch to alternative VLMs including Qwen2.5-VL-72B-Instruct, and Qwen2-VL-72B-Instruct~\cite{wang2024qwen2}. The initial temperature is set to 0.9 during caption generation.  to regenerate captions, which are then re-evaluated using the EVQAScore~\cite{liang2024evqascore}. Only those captions achieving a score above the quality threshold $\theta$ are retained.
This strategy leverages the diversity among VLMs—arising from differences in training data and architectural design—to improve caption quality for challenging clips. 
It is only when the number of failed attempts exceeds three that we resort to manual annotation to generate the final captions. This hybrid pipeline reduces the need for human intervention, thereby minimizing annotation cost while ensuring high-quality captioning at scale.

\subsubsection{Full Video Captioning}
Due to the length limitations of VLMs, directly generating captions for entire long videos often leads to incomplete or fragmented descriptions. To address this issue, we sequentially merge adjacent captions by aligning and fusing semantically related segments rather than naively concatenating all clip-level captions or processing them in an LLM beyond its sequence limit. The process is shown as follows:
\begin{equation}
\text{cap}_{\text{merged}} = 
p^{\text{rest}}_{\text{c}_1} + 
\text{LLM}\left( p^{\text{tail}}_{\text{c}_1} \parallel p^{\text{head}}_{\text{c}_2} \right) +
p^{\text{rest}}_{\text{c}_2}
\label{equation1}
\end{equation}

where $\text{LLM}$ denotes the rewriting process of the LLM. In our strategy, we connect the last paragraph of one clip with the first paragraph of the subsequent clip through a language model. During this process, the clip-level captions $ \text{cap}_c $ are sequentially aggregated to construct the final video-level caption $ \text{cap}_v $.

\subsubsection{Final Human Quality Assessment}
\label{final_check}
To assess the quality of the annotations in \textbf{LoVR}, we conducted a human evaluation study involving 25 participants. A total of 300 clips and 100 long videos were randomly sampled for evaluation. Each sample consisted of a video (or clip) and its associated caption. Participants were instructed to rate the relevance between the visual content and the caption on a 0–5 scale, where a score of 5 denotes perfect semantic alignment with no factual or descriptive errors.

The evaluation results indicate that the captions are of consistently high quality: the average rating across all samples was 4.3, and over 78\% of the samples received scores of 4 or 5. These findings validate the effectiveness of our annotation pipeline and demonstrate that the captions in \textbf{LoVR} accurately reflect the underlying video content.

\subsection{Dataset Statistics}\label{sec:dataset_statistics}

As shown in Table~\ref{table:benchmark-comparison}, the final dataset of \textbf{LoVR} consists of 467 videos and 40,804 fine-grained clips. Compared to conventional benchmarks such as MSR-VTT~\cite{xu2016msr} and MSVD~\cite{chen2011collecting}, our benchmark provides a larger scale of data, longer video durations, and more comprehensive captions at both the clip level and video level.

The annotation process employs a "VLM + Human" strategy, which achieves greater scalability than pure human annotation and higher quality than purely VLM-based methods. Furthermore, in the prompts designed for VLM annotation, the model is explicitly instructed to identify thematic information of the video, serving as contextual reference for video content.
The entire dataset is designated as the test set, and all data are used during experimental evaluation.

Figure~\ref{fig:clip_duration_distribution} illustrates key characteristics of the dataset. The figure (b) shows that the video captions in our dataset are notably longer and contain more detailed information. The figure (c) presents the duration distribution of clip-level videos. As the fundamental units of long videos, these clips cover short to medium lengths, and their combination forms the long video structure. This contrasts with existing benchmarks like MSVD~\cite{chen2011collecting}, MSR-VTT~\cite{xu2016msr}, and TGIF~\cite{tgif-cvpr2016}, which primarily focus on short video clips. A considerable proportion of medium-length videos (40–120 seconds) is included, accounting for approximately 15\% of the total collection. This distribution allows for a comprehensive evaluation of video retrieval models, testing their ability to comprehend brief visual semantics in short clips as well as complex temporal relationships in longer videos. The figure (d) from the left displays the distribution of clip caption lengths. It can be observed that the captions are highly detailed, with the majority exceeding 1,000 tokens, offering rich descriptions of the clip content.

\begin{table*}[t]
\caption{Baseline performance of video-to-text and clip-to-text retrieval on the \textbf{LoVR} benchmark.}
\centering
\resizebox{\textwidth}{!}{
\begin{tabular}{l|ccc|cc|ccc|cc}
\toprule
\multirow{2}{*}{\textbf{Model}} & \multicolumn{5}{c|}{\textbf{Video-to-Text Retrieval}} & \multicolumn{5}{c}{\textbf{Clip-to-Text Retrieval}} \\ 
\cmidrule{2-6} \cmidrule{7-11}
 & \textbf{R@1} & \textbf{R@5} & \textbf{R@10} & \textbf{Time w/ IO} & \textbf{Time w/o IO} & \textbf{R@1} & \textbf{R@5} & \textbf{R@10} & \textbf{Time w/ IO} & \textbf{Time w/o IO} \\
\midrule

\multicolumn{11}{c}{\textbf{Image Encoder}} \\ \midrule
CLIP& 16.70& 35.55& 47.11& 2544747& 1431709
& 18.29& 35.99& 44.76& 222270298& 125052299
\\
MetaCLIP-ViT-H-14 & 22.27& 46.68& 58.89& 1165648& 313637
& 28.32& 51.02& 60.52& 101813266& 27394590
\\
siglip-base-patch16-224 & 15.84& 36.40& 45.61& 1376746& 344674
& 21.02& 40.23& 49.50& 120251547& 30105548
\\
siglip2-base-patch16-224 & 7.49& 23.55& 31.48& 1336867& 269687
& 8.21& 18.51& 24.62& 116768361& 23555767
\\
EVA02-CLIP-B-16 & 16.27& 38.97& 48.61& 7491183& 5156938
& 23.25& 44.05& 53.83& 654315581& 450431558
\\
PE-Core-B16-224 & 18.42& 35.33& 43.68& 7460584& 4967311& 18.43& 36.00& 44.58& 651642897& 433868639\\
\midrule

\multicolumn{11}{c}{\textbf{Video Encoder}} \\ \midrule
VideoClip-XL & 29.12& 54.39& 66.38& 1074042& 29435
& 44.18& 69.95& 78.64& 93812003& 2571057
\\
VideoClip-XL-v2 & 25.05& 52.03& 64.03& 1073304& 29497& \textbf{48.54}& \textbf{73.82}& \textbf{81.75}& 93747515& 2576415\\
\midrule

\multicolumn{11}{c}{\textbf{Multimodal Embedding Model}} \\ \midrule
GME-Qwen2-VL& 8.35& 20.13& 28.91& 1680714676& 1598543574& 42.82& 66.03& 74.15& 19242308& 18301540\\
MM-Embed& 14.99& 37.26& 48.82& 2030535& 1168405
& 32.81& 53.55& 61.75& 177356626& 102054123
\\
LanguageBind-Video& \textbf{37.47}& \textbf{60.39}& \textbf{73.23}& 180170& 180170& 35.83& 61.12& 70.84& 15742371& 15742370\\
\bottomrule
\end{tabular}
}
\label{table:retrieval_table2_new}
\end{table*}

\section{Experiments}
\label{sec:experiments}

\subsection{Experimental Settings}

\subsubsection{Evaluation Metrics.}
We utilize the following two metrics when conducting the quantitative evaluation.
\paragraph{Retrieval Accuracy} This metric is measured using Recall@K, which quantifies the proportion of true relevant items among the top-K retrieved results and reflects the recall capability of the retrieval model. For video retrieval with image-based models, we adopt the \textit{average feature pooling} strategy, where frame-level embeddings are averaged into a single video-level representation before computing similarity with the caption.

\paragraph{Retrieval Time} This metric is measured in ms, under two configurations: (1) \textit{w/ I/O}: End-to-end time including disk read/write and network transmission, reflecting real-world deployment bottlenecks. (2) \textit{w/o I/O}: Pure computation time excluding I/O operations, such as vector similarity calculation and result ranking, used to assess algorithmic optimization potential.

\subsubsection{Baselines}
In order to fully evaluate our benchmark, we select three categories of widely used retrieval models. The following are the baselines used:

\paragraph{Image Encoders.}
\text{CLIP}~\cite{radford2021learning} is a foundational vision-language model that learns a shared embedding space for images and text through contrastive pre-training.
\text{MetaCLIP-ViT-H-14}~\cite{xu2023demystifying} is a large-scale CLIP variant trained on a vast dataset, utilizing a Vision Transformer (ViT-H/14) backbone for enhanced visual representation.
\text{SigLIP-base-patch16-224}~\cite{zhai2023sigmoid} improves upon CLIP by employing a sigmoid-based loss function that operates on individual image-text pairs.
\text{SigLIP2-base-patch16-224}~\cite{tschannen2025siglip} is an advanced iteration of SigLIP that integrates multiple pretraining techniques into a unified recipe.
\text{EVA02-CLIP-B-16}~\cite{fang2024eva} leverages the robustly optimized EVA-02 Vision Transformer as its image encoder, offering strong performance in vision-language tasks.
\text{PE-Core-B16-224}~\cite{bolya2025perception} is a competitive perception encoder trained with a robust vision-language learning objective, designed for high performance in image and video understanding tasks.

\paragraph{Video Encoders.}
\text{VideoCLIP-XL}~\cite{wang2024videoclip} is specifically designed for long video understanding, leveraging a large-scale dataset of video-long description pairs to enhance its comprehension of extended temporal content.
\text{VideoCLIP-XL-v2}~\cite{wang2024videoclip} is an optimized successor to VideoCLIP-XL, trained with additional data and refined techniques to further boost its capabilities in long-form video-text retrieval.

\paragraph{Multimodal Embedding Models.}
\text{GME-Qwen2-VL}~\cite{zhang2024gme} is a unified multimodal embedding model built upon the Qwen2-VL multimodal large language model, capable of generating joint embeddings for text, images, and image-text pairs.
\text{MM-Embed}~\cite{lin2024mm} is a universal multimodal retrieval model based on a multimodal large language model (MLLM) with a bi-encoder architecture, designed for efficient and broad-spectrum cross-modal search.
\text{LanguageBind-Video}~\cite{zhu2023languagebind} adopts a language-centric approach to align video features with other modalities in a shared semantic space.

\begin{figure}[h]
    \centering
    \includegraphics[width=0.95\columnwidth]{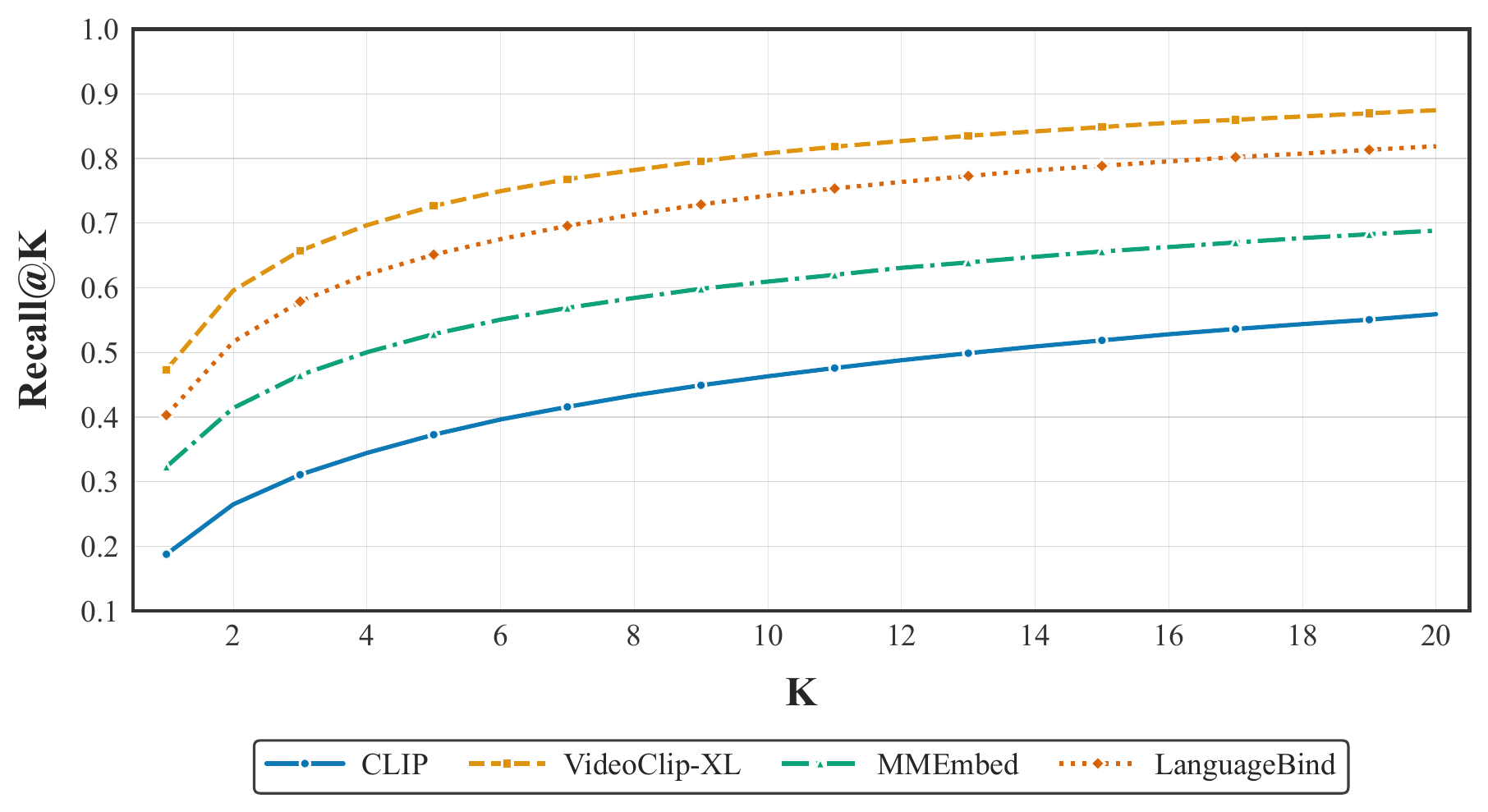}
    \caption{Recall@K performance comparison across baseline models on LoVR clip-level retrieval.}
\label{fig:clip_duration_distribution}
\end{figure}

\begin{table*}[t]
\centering
\caption{Long video retrieval performance under different frame counts across retrieval tasks.}
\label{tab:frame_count_all}
\resizebox{\linewidth}{!}{
\begin{tabular}{l|c|ccc|ccc|ccc|ccc}
\toprule
\multirow{2}{*}{\textbf{Model}} & \multirow{2}{*}{\textbf{Frames}} 
& \multicolumn{3}{c|}{\textbf{Text-to-Clip}} 
& \multicolumn{3}{c|}{\textbf{Text-to-Video}} 
& \multicolumn{3}{c|}{\textbf{Clip-to-Text}} 
& \multicolumn{3}{c}{\textbf{Video-to-Text}} \\
\cmidrule(lr){3-5}\cmidrule(lr){6-8}\cmidrule(lr){9-11}\cmidrule(lr){12-14}
& & \textbf{R@1} & \textbf{R@5} & \textbf{R@10}
  & \textbf{R@1} & \textbf{R@5} & \textbf{R@10}
  & \textbf{R@1} & \textbf{R@5} & \textbf{R@10}
  & \textbf{R@1} & \textbf{R@5} & \textbf{R@10} \\
\midrule
\multirow{3}{*}{Clip} 
& 10 & 18.82& 37.50& 46.29& 23.34& 43.90& 54.39& 18.16& 35.82& 44.64& 16.49& 35.76& 46.90\\
& 20 & 18.94& 37.50& 46.49& 23.13& 43.90& 54.39& 18.20& 35.82& 44.75& 16.49& 35.76& 46.68\\
& 30 & 18.78& 37.28& 46.31& 23.34& 43.90& 54.39& 18.20& 35.99& 44.76& 16.70& 35.55& 47.11\\
\midrule
\multirow{3}{*}{VideoCLIP-XL} 
& 10 & \textbf{49.06} & \textbf{73.69} & \textbf{81.66} & 30.62& 54.60& 64.45& \textbf{45.36} & \textbf{70.96} & \textbf{79.41} & \textbf{30.84} & 54.39& \textbf{66.81} \\
& 20 & 48.31& 73.34& 81.35& \textbf{30.84} & \textbf{55.46} & \textbf{67.02} & 44.66& 70.67& 78.86& 29.76& \textbf{54.60} & 66.17\\
& 30 & 47.33& 72.67& 80.81& \textbf{30.84} & 54.82& 65.74& 44.18& 69.95& 78.64& 29.12& 54.39& 66.38\\
\midrule
\multirow{3}{*}{MM-Embed} 
& 10 & 32.64& 53.21& 61.35& 14.13& 31.69& 40.90& 32.95& 53.69& 61.80& 14.99& 37.47& 49.25\\
& 20 & 32.36& 53.04& 61.20& 14.13& 31.69& 40.69& 32.81& 53.71& 61.76& 14.99& 37.47& 49.25\\
& 30 & 32.32& 52.84& 60.96& 14.13& 31.05& 40.69& 32.81& 53.55& 61.75& 14.99& 37.36& 48.82\\
\bottomrule
\end{tabular}
}
\end{table*}

\subsection{Baseline Performance on Video Retrieval and Clip Retrieval}
We benchmark a comprehensive suite of models on the \textbf{LoVR} dataset across two fundamental retrieval tasks: text-to-video/clip and video/clip-to-text (see Tables~\ref{table:retrieval_table1_new} and~\ref{table:retrieval_table2_new}).
Overall, \textbf{LanguageBind-Video} emerges as the most effective model among all evaluated baselines, achieving the highest performance in both text-to-video (42.6\% R@1) and video-to-text (37.5\% R@1) retrieval. Its clip-level performance is also strong, reaching 40.3\% R@1 in text-to-clip and 35.8\% R@1 in clip-to-text retrieval.
In contrast, image-based encoders such as \textbf{CLIP} and \textbf{MetaCLIP-ViT-H-14} show limited scalability when handling long-form videos. For instance, CLIP achieves only 23.3\% R@1 in text-to-video retrieval and 16.7\% R@1 in video-to-text retrieval, demonstrating poor temporal modeling capacity.
\textbf{VideoClip-XL-v2}, as a dedicated video encoder, exhibits solid performance in short-clip retrieval (55.3\% R@1 in text-to-clip and 48.5\% R@1 in clip-to-text), but its performance on full-length video retrieval remains moderate (29.9\% R@1 in text-to-video and 25.1\% R@1 in video-to-text).
Among multimodal embedding models, \textbf{GME-Qwen2-VL} and \textbf{MM-Embed} demonstrate good clip-level recall (53.5\% and 32.3\% R@1 in text-to-clip, respectively), yet both suffer a substantial drop on long videos, indicating that multimodal embeddings trained primarily on images struggle to generalize to extended temporal contexts.

These results collectively highlight the inherent difficulty of the \textbf{LoVR} benchmark. Its long temporal dependencies, rich semantic variation, and stringent cross-modal alignment requirements make it a rigorous and realistic testbed for evaluating scalable long video–text retrieval systems.

\subsection{Comparative Analysis of Recall@K for Baseline Models}

Figure~\ref{fig:clip_duration_distribution} presents the Recall@K performance of representative baselines on the clip-level retrieval task. \textbf{VideoClip-XL} achieves the strongest overall performance, with Recall@1 of 47.3\% and Recall@20 of 87.5\%, demonstrating superior temporal modeling and cross-modal alignment. \textbf{LanguageBind} follows closely, reaching 40.3\% at Recall@1 and 81.9\% at Recall@20, while \textbf{MM-Embed} remains moderate, with Recall@1 of 32.3\% and Recall@20 below 69.0\%. In contrast, \textbf{CLIP} performs weakest, starting at only 18.8\% and reaching just 55.9\% even when $K=20$.

Notably, despite steady improvement across all models as $K$ increases, even the best-performing model fails to achieve perfect recall at large $K$. This plateau effect underscores the intrinsic difficulty of the \textbf{LoVR} benchmark: it requires precise long-range temporal reasoning and robust semantic grounding to retrieve relevant content among diverse, real-world video scenes. Consequently, LoVR serves as a rigorous and discriminative benchmark for advancing the next generation of video–language retrieval systems.

\subsection{Impact of Frame Count in Video Embedding on Retrieval Performance}
As shown in Table~\ref{tab:frame_count_all}, varying the number of sampled frames leads to only marginal differences in retrieval accuracy. While using more frames provides additional visual information, the improvements are inconsistent and often negligible across tasks. For example, CLIP shows virtually no improvement when increasing frames from 10 to 30 across all tasks, with performance remaining nearly flat. MM-Embed exhibits similar behavior with minimal variations. Even VideoCLIP-XL—though achieving the best overall performance—demonstrates mixed results: while Text-to-Video tasks show slight improvements with more frames, Text-to-Clip and Video-to-Text tasks actually perform best with 10 frames, suggesting that simply adding more frames may introduce noise rather than useful information.These results underscore the inherent difficulty of the LoVR benchmark. Despite employing advanced visual feature extractors, the retrieval performance remains far from satisfactory, particularly evident in the large performance gaps between different model architectures. The benchmark therefore presents a demanding setting that cannot be solved by naive frame sampling strategies alone, but instead calls for more sophisticated algorithms and models capable of robust long-video understanding and cross-modal alignment.

\section{Conclusion}
In this work, we present \textbf{LoVR}, a novel benchmark tailored for long video-text retrieval, addressing key limitations in existing datasets such as short video duration, low-quality captions, and insufficient annotation granularity. LoVR features 467 long videos and over 40,000 high-quality, fine-grained clips, supported by a scalable and effective caption generation pipeline that combines VLM-based generation, automatic quality assessment, and iterative refinement. Additionally, we propose a semantic fusion method to produce coherent full-video captions, preserving crucial contextual information. Extensive evaluations using state-of-the-art embedding models demonstrate the challenges posed by LoVR and highlight the limitations of current retrieval methods. We believe that LoVR will serve as a valuable resource for advancing research in multimodal video understanding and retrieval.


\balance
\bibliographystyle{ACM-Reference-Format}
\bibliography{main1}

\appendix
\newpage
\section{Details of LoVR Construction}
\subsection{Implementation Details.}
\label{app:implementation_details}

The inter-frame difference scores shown in Figure \ref{fig:video_type_comparison} are calculated by computing the differences between the average pixel values of consecutive frames. To ensure a robust evaluation, we run each quantitative experiment three times and take the average. In the baseline approach, all encoder models first vectorize the videos and the query captions into embeddings. Then, the cosine similarity between each video feature and the query caption feature is computed and used for ranking.

\subsection{Computational Costs.}
\label{Computational_Costs}
The caption generation process using Qwen2.5-VL-72B-Instruct~\cite{bai2025qwen2} requires approximately 720 GPU hours. Evaluation of the generated captions with the EVQAScore~\cite{liang2024evqascore} model adds approximately 100 GPU hours to the total computational load. Consequently, constructing \textbf{LoVR} incurs a total computational cost of roughly 820 GPU hours on NVIDIA H800 GPUs. All experiments are conducted on NVIDIA H800 GPUs.

\section{Details of Human Quality Assessment}
\label{human_assessment}

In order to ensure the overall quality and reliability of the video-caption pairs in our dataset, we conducted multiple human evaluation processes. This included: (1) \textit{video filter assessment}, (2) \textit{threshold determination process}, (3) \textit{human annotation}, and (4) \textit{final quality assessment}.

\subsection{High-scene-dynamic Video Filter Assessment}
\label{app:video_filter_assessment}
This step is conducted after filtering out the high-scene-dynamic videos (Section \ref{sec:video_filtering}).
We recruited 10 undergraduate student volunteers with computer science backgrounds to manually review all videos that had passed the filter. The goal was to ensure that only high-dynamic videos were retained in the final dataset. Here, high-dynamic videos are defined as those containing significant scene changes, camera transitions, moving characters or objects, and diverse semantic content.

To enhance the reliability of this filtering stage, each video was independently reviewed by at least two annotators. If both agreed that a video lacked sufficient visual dynamics, it was excluded from the dataset.
The entire review process took approximately 2 days, with each participant spending an average of 4 hours on the assigned tasks. A total compensation of ¥3,000 (¥ is the currency symbol for the Chinese yuan (RMB)) was provided to the annotators. As a result of this manual verification, we confirmed and retained 467 high-quality long videos in the final dataset.

\subsection{Threshold Determination Process}
This step is conducted after scoring the captions generated by VLMs (Section \ref{caption_generation}).
Each video was assessed by three independent annotators who were tasked with identifying inconsistencies between the captions and the video content. These inconsistencies included textual errors, redundant or missing information, and unaddressed scene transitions. In particular, if a clear scene change occurred in the video (lasting more than 3 seconds) but was not described in the caption, annotators were required to specify its exact duration. Additionally, they evaluated whether the overall theme of the caption significantly deviated from that of the video.

A total of 500 videos were selected for annotation, with each video associated with one clip-level and one video-level caption, resulting in 1,500 annotations in total. Video durations ranged from 3 to 60 seconds. To ensure annotation consistency and quality, six annotators with relevant backgrounds were recruited, each responsible for annotating approximately 250 samples. On average, each annotator spent around 300 minutes on the task. The total project budget was approximately ¥1,200.
By analyzing the trends in the manual scores, we were able to preliminarily determine a reasonable threshold for caption quality filtering, which serves as a foundation for subsequent automated evaluation and data curation.

\subsection{Human Annotation}

This step is conducted to improve the quality of captions getting low scores (Section \ref{caption_generation}).
To ensure the high quality and consistency of the textual descriptions used in the video-text retrieval tasks, we conducted a detailed annotation process involving a total of 1,171 videos, each ranging from 3 to 60 seconds in duration. The annotation effort was distributed among four annotators, with each annotator reviewing approximately 293 videos. Each annotator received a base payment of ¥250, and those achieving an accuracy rate above 90\% on randomly sampled checks (30 per annotator) received an additional bonus of ¥30, leading to a maximum compensation of ¥280 per person.

The annotation guidelines primarily focused on two categories of issues: \textit{contextual content errors} and \textit{topic description discrepancies}. Specifically, contextual content errors were further divided into three types:
\begin{itemize}
    \item \textbf{A}: Textual expression errors.
    \item \textbf{B}: Redundant text.
    \item \textbf{C}: Missing key segments, such as unmentioned transitions between clearly distinguishable shots lasting more than 3 seconds.
\end{itemize}
Topic description discrepancies were categorized as type \textbf{D}, indicating that the overall textual description was clearly inconsistent with the video content.

During annotation, if any error was identified within a sentence, the entire sentence was marked using delimiters `{{{' and `}}}'. For type C errors (i.e., missing content), the placeholder `{{{ }}}' was inserted at the appropriate location where content should be added.

In terms of annotation format, annotators were required to follow a structured labeling protocol when filling out the annotation table:
\begin{itemize}
    \item In the ``Type'' column, the corresponding error types (A, B, C, or D) were recorded. When multiple error types occurred within the same sentence, they were separated by `\#`, e.g., A\#B\#C or B\#D.
    \item In the ``Correction'' column, the corrected version of the full sentence was provided. Regardless of how many error types were present in a sentence, only one final corrected version was written, although all relevant error types were listed in the ``Type'' column.
    \item For type B errors (redundant content), if an entire sentence needed to be deleted, the value `0` was entered in the ``Correction'' column. If only part of the sentence was redundant, the revised sentence with the unnecessary parts removed was recorded.
    \item Annotators were also required to count the number of content-related errors (A+B+C) and topic-level errors (D) for each video and record these statistics in designated fields.
\end{itemize}

Throughout the annotation workflow, we emphasized consistency in annotation standards to ensure that all corrected texts remained semantically fluent, logically coherent, and accurately reflected the visual content of the videos after replacing the original captions. This annotation mechanism not only improved the overall quality of the dataset but also provided a reliable foundation for subsequent model training and evaluation.
\begin{figure}[ht]
    \centering
    \includegraphics[width=\linewidth]{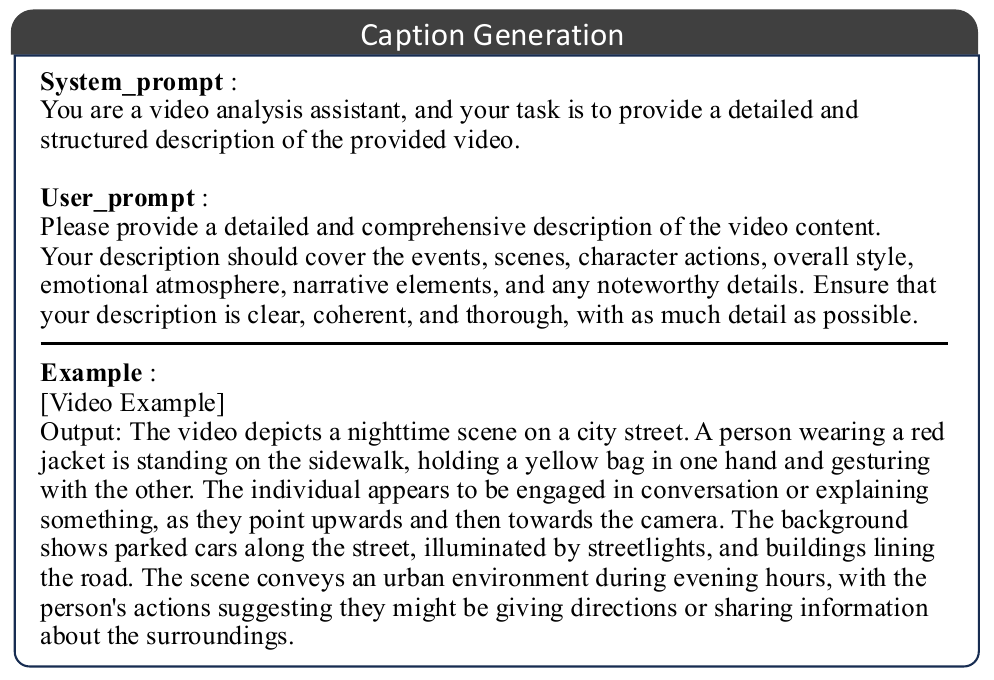}
    \caption{Prompt used for clip caption generation.}
\label{fig:caption_generation_prompt}
\end{figure}

\begin{figure}[ht]
    \centering
    \includegraphics[width=\linewidth]{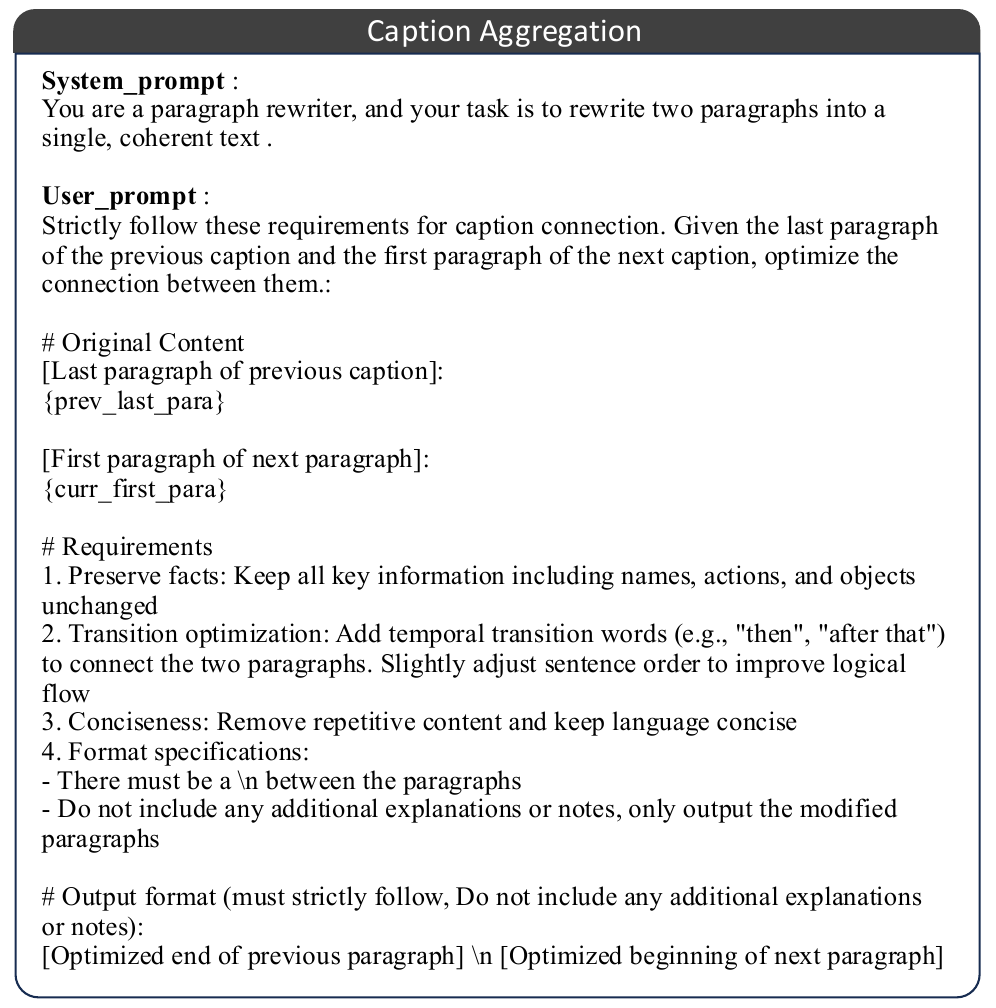}
    \caption{Prompt used for caption aggregation.}
\label{fig:caption_aggragation_prompt}
\end{figure}

\begin{figure}[ht]
    \centering
    \includegraphics[width=\linewidth]{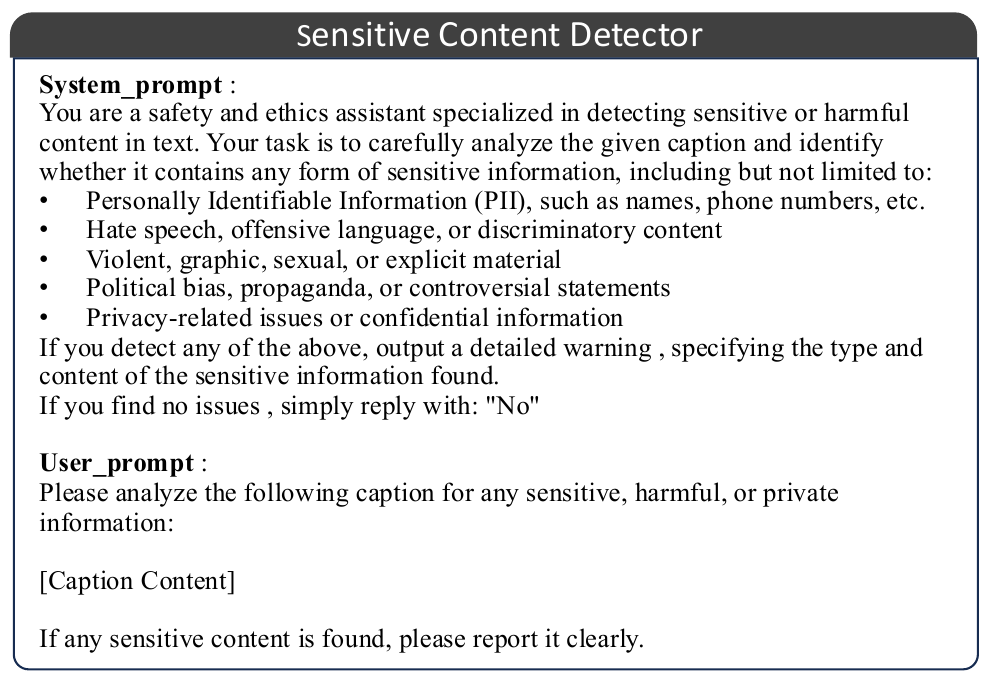}
    \caption{Prompt used for sensitive content detection.}
\label{fig:sensitive_prompt}
\end{figure}
\subsection{Final Quality Assessment}
As shown in Section~\ref{final_check}, to evaluate the semantic alignment between videos and their corresponding captions in the final dataset (\textbf{LoVR}), we conducted a comprehensive human evaluation involving 25 carefully selected participants. All participants were students with backgrounds in computer science.

Prior to the evaluation, all participants received a detailed instruction manual and completed a short training session using sample videos to standardize their understanding of the scoring criteria.
A total of 300 clips and 100 long videos were randomly sampled from the dataset for evaluation. Each sample consisted of a video (or clip) paired with its corresponding caption. Participants were asked to rate the semantic alignment between the visual content and the caption on a 6-point Likert scale (0–5), where:
\begin{itemize}
    \item \textbf{5}: Perfect match — no factual or descriptive errors.
    \item \textbf{4}: Minor issues but overall accurate.
    \item \textbf{3}: Moderate mismatch or ambiguity.
    \item \textbf{2}: Significant inaccuracies.
    \item \textbf{1}: Nearly irrelevant.
    \item \textbf{0}: Completely unrelated.
\end{itemize}

Each participant was assigned approximately 16 clips and 4 long videos. On average, about 180 seconds were spent per clip, including time for reading instructions and providing ratings. The entire annotation task took roughly one hour per participant. Participants were compensated at a rate of ¥60 per hour.

\section{LLM Prompt}

Figure~\ref{fig:caption_generation_prompt} shows the prompt used for clip caption generation, Figure~\ref{fig:caption_aggragation_prompt} shows the prompt used for caption aggregation, and Figure~\ref{fig:caption_aggragation_prompt} shows the prompt used for sensitive content detection of captions.

\begin{figure}[h]
    \centering
    \hspace{0.03\textwidth} 
    \begin{minipage}{0.47\textwidth}
        \centering
        \includegraphics[width=0.99\textwidth]{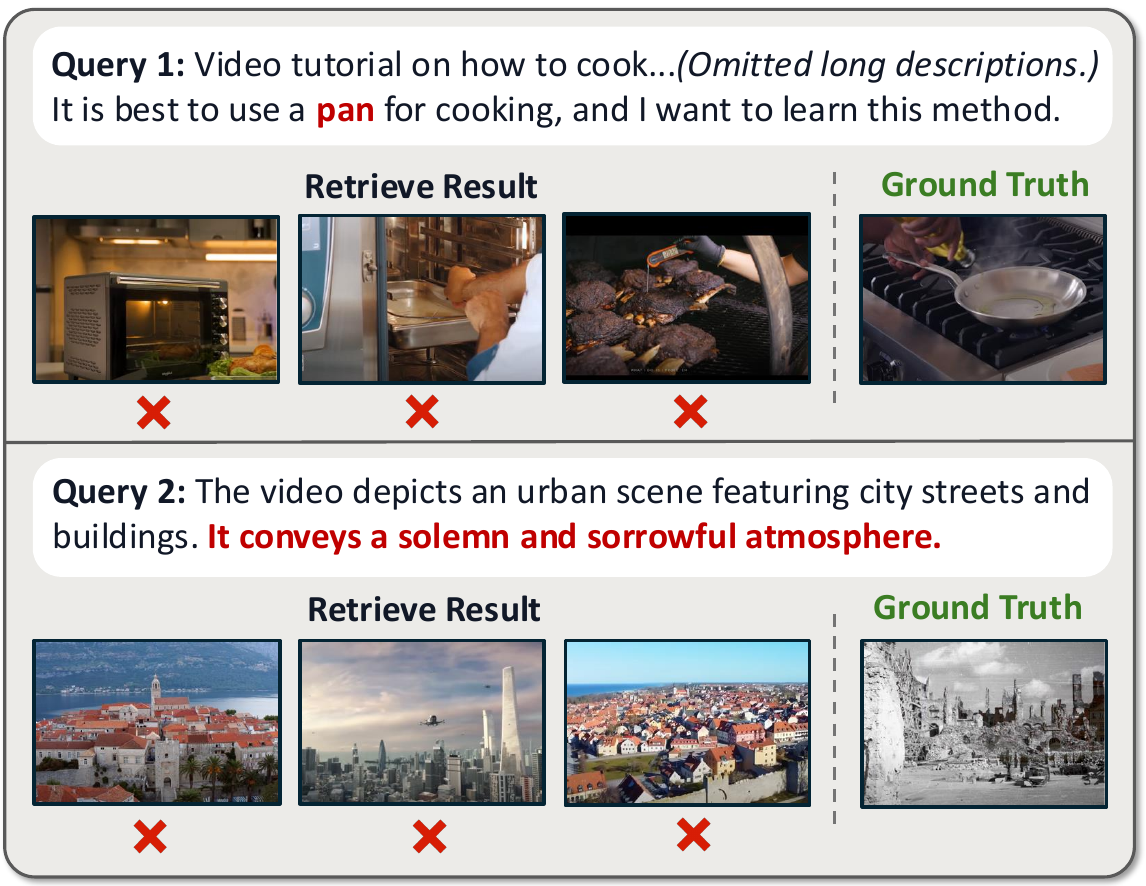}
        \caption{Two illustrative retrieval cases are provided, where the key retrieval targets are highlighted in red.}
        \label{fig:case_study}
    \end{minipage}
\end{figure}

\section{Case Study}

As shown in Figure~\ref{fig:case_study}, two example failure cases are presented. In Query 1, due to the length of the input text, we omit the content of the middle portion. It can be observed that under long-query conditions, the model fails to process the entire context, resulting in the neglect of information at the end of the query. This leads to the omission of critical details — for instance, the key element "pan" in this case is ignored. In Query 2, the model demonstrates limitations in capturing theme and atmosphere information. Although four videos related to urban city scenes are retrieved, only the first one accurately reflects the required `solemn and sorrowful atmosphere.' The remaining results fail to incorporate this essential emotional aspect, leading to incorrect retrievals. These cases highlight the importance of accurate long context understanding in long video retrieval tasks.

\end{document}